\documentclass{article}

\usepackage[preprint]{neurips_2024}

\usepackage[utf8]{inputenc} 
\usepackage[T1]{fontenc}    
\usepackage{microtype}      
\usepackage{setspace}
\usepackage{hyperref}       
\usepackage{url}            
\usepackage{booktabs}       
\usepackage{amsfonts}       
\usepackage{nicefrac}       
\usepackage{xcolor}         
\usepackage{colortbl}
\usepackage{amsmath}
\usepackage{amssymb}
\usepackage{xkcdcolors}
\usepackage{multirow}
\usepackage{graphicx}
\usepackage{capt-of}
\usepackage{wrapfig}
\newcommand{\eg}{\emph{e.g.}}
\newcommand{\ie}{\emph{i.e.}}
\newcommand{\etc}{\emph{etc.}}

\title{AnyLogo: Symbiotic Subject-Driven Diffusion System with Gemini Status}

\author{%
  Jinghao Zhang$^{1}$\thanks{This work was done during an internship at Alibaba Group. This work was supported by Alibaba Group through Alibaba Research Intern Program.} \quad Wen Qian$^{2}$ \quad Hao Luo$^{2,3}$ \quad Fan Wang$^{2}$ \quad Feng Zhao$^{1}$\thanks{Corresponding author.}\\
  \textsuperscript{1}University of Science and Technology of China, \\
  \textsuperscript{2}DAMO Academy, Alibaba Group
  \quad \textsuperscript{3}Hupan Lab, Zhejiang Province
}

\begin{document}

\maketitle

\begin{abstract}
  Diffusion models have made compelling progress on facilitating high-throughput daily production.
  Nevertheless, the appealing customized requirements are remain suffered from instance-level finetuning for authentic fidelity. 
  Prior zero-shot customization works achieve the semantic consistence through the condensed injection of identity features, while addressing detailed low-level signatures through complex model configurations and subject-specific fabrications, which significantly break the statistical coherence within the overall system and limit the applicability across various scenarios.
  To facilitate the generic signature concentration with rectified efficiency,
  we present \textbf{AnyLogo}, a zero-shot region customizer with remarkable detail consistency, 
  building upon the symbiotic diffusion system with eliminated cumbersome designs.
  Streamlined as vanilla image generation,
  we discern that the rigorous signature extraction and creative content generation are promisingly compatible and can be systematically recycled within a single denoising model.
  In place of the external configurations, the gemini status of the denoising model promote the reinforced subject transmission efficiency and disentangled semantic-signature space with continuous signature decoration. 
  Moreover, the sparse recycling paradigm is adopted to prevent the duplicated risk with compressed transmission quota for diversified signature stimulation.
  Extensive experiments on constructed logo-level benchmarks demonstrate the effectiveness and practicability of our methods.
\end{abstract}

\section{Introduction}
\label{sec:intro}
Diffusion models have demonstrated impressive capabilities in creative content generation, such as marvelous image generation~\cite{rombach2022high,podell2023sdxl,ramesh2022hierarchical,saharia2022photorealistic}, image manipulation~
\cite{brooks2023instructpix2pix,kawar2023imagic,hertz2022prompt,hu2024instruct}, video generation~\cite{videoworldsimulators2024,bar2024lumiere,esser2023structure,long2024videodrafter}, audio synchronized~\cite{zhang2024audio,he2024co,novack2024ditto,tan2024edtalk}, \etc, and profoundly impact the practice of the public daily production. 
Large volume of the generative requests necessitate the expeditious response, while the appealing customized aspirations are remain suffered from instance-level finetuning for authentic fidelity. 
Prosperously, recent efforts have been witnessed toward the zero-shot customization with single forward pass, which achieve the semantic consistency through the condensed injection of identity features,
applications involving the object-level manipulation~\cite{yang2023paint,song2023objectstitch,pan2024locate,ye2023ip}, facial fidelity generation~\cite{li2023photomaker,wang2024stableidentity,yan2023facestudio,chen2023dreamidentity}, \textit{etc},
while the meticulous low-level signatures are less concentrated.

\definecolor{c1}{HTML}{177cb0}
\newcommand{\symbotic}{\color{orange}}
\begin{figure*}[t]\small
\begin{center}
   \includegraphics[width=1\linewidth]{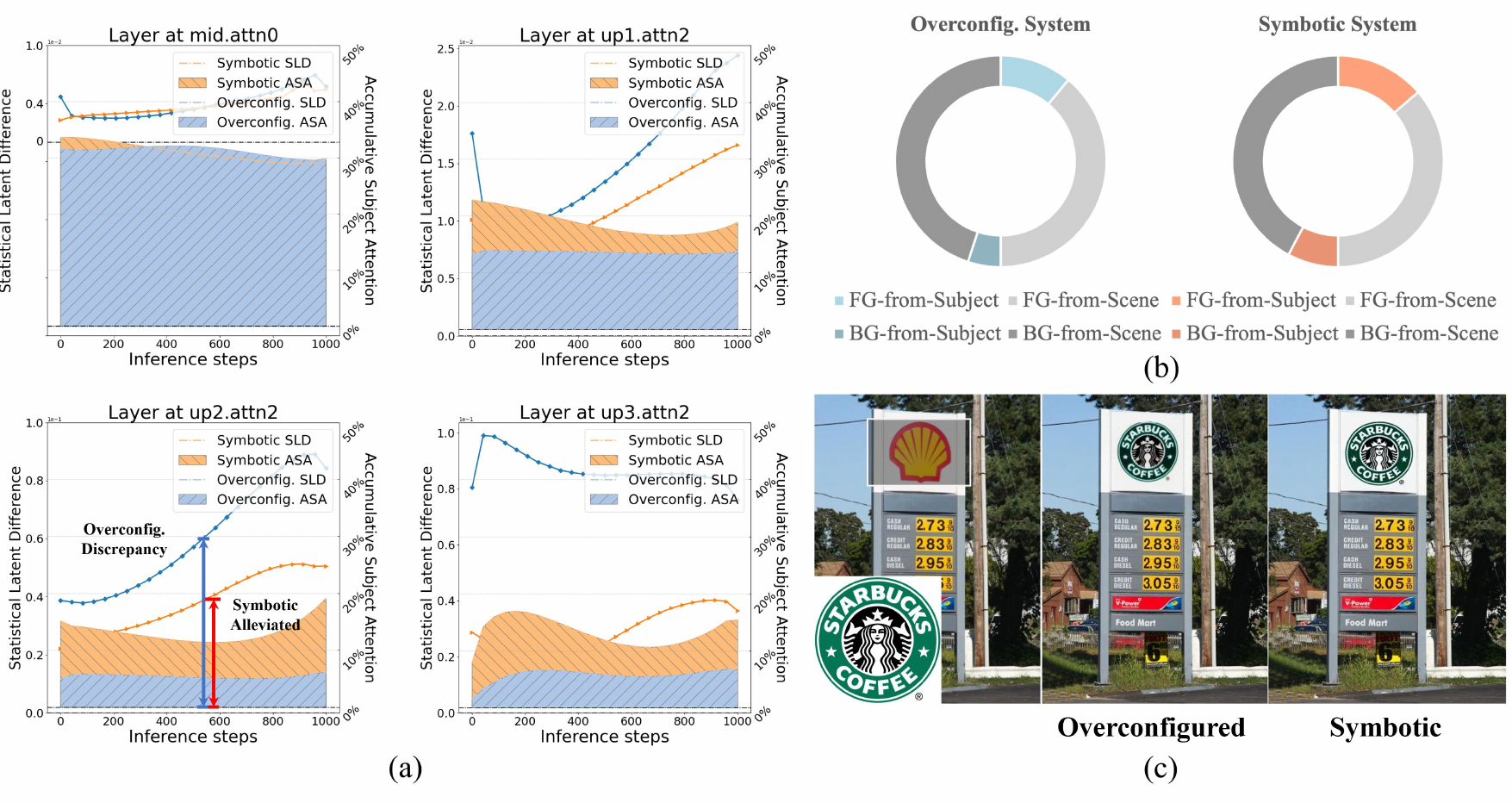}
\end{center}
\vspace{-1.6em}
   \caption{The comparison of the \textcolor{c1}{overconfigured system} and the {\symbotic symbotic system} in transmission efficiency. (a) The proportion of the accumulative subject attention (ASA) and the transmitted statistic latent difference (SLD) at four distributed self-attention layers in the denoising model, where the SLD is computed between the transmitted subject latents and the corresponding denoising latents. The symbotic system raises the increasing transmission efficiency with deeper model operations. (b) The comprehensive attention analysis accumulated along the model layers and the denoising steps. Both the customized region (light) and the background area (dark) benefit the boosted subject expertise from the symbotic system. (c) Visual comparison results of two subject-driven diffusion systems. Detailed calculation process and visual illustration of two systems are provided in Appendix~\ref{sec: symbotic}.}
\label{fig:stas}
\vspace{-1em}
\end{figure*}

In this regard, we may entertain that \textit{are we really need the low-level signatures in daily customization}. 
Beyond the semantic recognizable similarities, there are numerous fidelity-disciplined sceneries such as text glyph generation~\cite{tuo2023anytext,chen2024textdiffuser,zhang2024brush}, image character animation~\cite{wang2023disco,hu2023animate,xu2023magicanimate,ma2024follow}, virtual tryon~\cite{seyfioglu2024diffuse,zang2024product,zhang2024acdg}, \etc, that strive for the extraordinary signature consistency with lower vision tolerance.
Moreover, the brand logo customization drastically involves the copyright license,
which overwhelming the moderate semantic injection insufficient for the signature-consistent customization, as shown in Fig.~\ref{fig:visual}.
Current practices incorporate the auxiliary configured ControlNet~\cite{tuo2023anytext,zang2024product,chen2023anydoor} and ReferenceNet~\cite{hu2023animate,xu2023magicanimate,wang2024instantid} for reinforced detail enhancement through progressive residual complement or hierarchical mutual spatial attention.
However, the complex model overconfigurations significantly break the statistical coherence within the overall system, as shown in Fig.~\ref{fig:stas} (a), resulting in suboptimal signature transmission efficiency with declined proportion of the accumulated subject attention score, which we referred as \textit{overconfigured system}.
Besides, there are also specialized attempts involving the utilization of the OCR model in precise text generation~\cite{tuo2023anytext} and the posture steered warping operation in tryon application~\cite{zang2024product,zhang2024acdg}, however, the subject-specific fabrications considerably narrow the applicable scenario potential.
In light of this, it is prospective to facilitate the generic signature concentration in daily customization with rectified transmission efficiency.

To this end, we present AnyLogo, a zero-shot region customizer with remarkable low-level signatures consistency, 
proficient in diversified graphic patterns and text glyphs.
Streamlined as vanilla image generation, 
AnyLogo is built upon the symbiotic diffusion system with economic model recycling policy, 
where we discern that \textbf{the rigorous signature extraction and creative content generation can be systematically recycled within a single denoising network}. 
The symbiotic diffusion system manifest the promisingly compatible generation capability with several peculiarities:
\begin{itemize}
    \item In place of the external configuration, the gemini status of the denoising network, \ie, the signature extraction and content generation largely alleviate the statistic coherence of the system owing to the self-delivery, yielding efficient signature transmission with reinforced subject attention, as shown in Fig.~\ref{fig:stas} (a) and (b). 
    And the efficiency raised by symbotic system manifests the increased momentum as more operations within the denoising model.
    \item The low-level signatures derived from the symbiotic system is highly semantic-independent, where the native diversified generation capability is conserved with blocked signature delivering, which is disparate from the collapsed status in overconfigured systems, as shown in Fig~\ref{fig:interpo}. Moreover, the symbiotic system enjoy the continuous signature decoration space.
\end{itemize}
Apart from that, the overloaded signatures generically incur the potential duplicated risk. 
Preceding works introduce various compressed signals such as 
landmark representation~\cite{wang2024instantid},
high-frequency map~\cite{zang2024product,chen2023anydoor} for signature delivering, 
which is unaccommodated to the symbotic system owing to the altered signal states.
In consequence, we adopt the sparse recycling paradigm with randomly discarding the self-delivered signatures for trimmed transmission quota, stimulating diversified hyper-representations of the signature with scene harmonization. 
Additionally, we show that Anylogo also supports the diversified highlight of the specified subject area with sensible background translation.

To comprehensively evaluate our method, a logo-level customization benchmark is constructed, involving $\sim$1k high-quality pairs with rich textures, collecting from the open source wild logo detection datasets, tryon datasets with brand annotation, and text glyph datasets.
Extensive experiments on constructed benchmarks demonstrate the effectiveness and practicability of our methods.

\section{Related work}
\label{Related}
\subsection{Prompt-driven Image Region Manipulation}
Creative and convenient image region manipulation has attracted increasing number of people to exercise in their daily production, accredited to the remarkable progress of the prevailing generative model.
Basically, the user practice can be categorized into the following three paradigms.
The text-driven image region manipulation~\cite{nichol2021glide,avrahami2022blended,zhuang2023task,manukyan2023hd} operates the candidate region with single text prompt, which specifies the desired attribute or object in principle, result in semantic aligned appearance. 
Despite the flexibility, the precise control manifests to be imperative.
The image-driven image region manipulation
~\cite{yang2023paint,song2023objectstitch} transports the image prompt to the candidate region, delivering the concrete manipulative intention, result in content preservation and realistic outlook.
Traditional image composition methods have investigated such aspiration for a long while, including image harmonization~\cite{xue2022dccf,chen2023hierarchical}, image blending~\cite{zhang2020deep,wu2019gp}, and geometric correction~\cite{azadi2020compositional,lin2018st}.
However, the intricate subbranches perplex the feasible pipelines, and the low-level manipulation is more concentrated with restricted semantic conformity.
Recent diffusion models~\cite{rombach2022high,podell2023sdxl} have shown impressive image composition capability, which incorporate the image prompt into the denoising process through condensed encoding embedding injection~\cite{yang2023paint,song2023objectstitch,ye2023ip,yu2023inpaint}.
While the details are further complemented with low-level feature interaction~\cite{chen2023anydoor,lu2023tf,wang2024primecomposer,zhang2023paste}.
Toward the synergic, the multimodal-driven image manipulation~\cite{li2024unimo,hu2024instruct} are further investigated with interleaved text and image prompt, fortified with multimodal large-language models for diversified token injection.

\subsection{ID-preserved Image Generation}
ID-preserved image generation is widely requisite in applications. Prior works on customized concept learning~\cite{ruiz2023dreambooth,gal2022image,gu2024mix,kumari2023multi,liu2023cones} generate subject-relevant images with arbitrary text prompt in few of user provided images. However, the optimization typically involves the intensive instant-level finetuning, which is cumbersome for large-scale deploy.
To this end, recent efforts on facial identity preservation~\cite{wang2024instantid,li2023photomaker,wang2024stableidentity,yan2023facestudio,chen2023dreamidentity} accomplish the customization with a single forward pass, depending on the injection of the identity feature extracted from CLIP or facial model, result in the semantic recognizable fidelity and flexible text controllability.
In the field of human animation, the image-to-video methods~\cite{xu2023magicanimate,wang2023disco,hu2023animate,long2024videodrafter,feng2023dreamoving} generate the reference-based video sequences following the motion signals, and relies heavily on the identity consistency. The common practice predefine the configured ControlNet or ReferenceNet to retain precious subject details with hierarchical representation interaction, result in impressive id preservation.
Additionally, there are also attempts toward the training-free customization~\cite{lu2023tf,wang2024primecomposer,cao2023masactrl,xu2023inversion} through multi-branch attention manipulation with paralled reconstructive diffusion processes, albeit the flexibility, they reckon on the pretrained model with confined fidelity and concentrate on the original text-driven manipulation.

\section{Method}
\label{method}
In this section, we provide a brief background of the text-to-image diffusion models and current subject-driven customization practices in Sec.~\ref{sec:Preliminary}.
Then, we introduce our symbiotic diffusion system built with model recycling policy in Sec.~\ref{sec:recycling}.
The sparse recycling with compressed transmission quota is presented in Sec.~\ref{sec:sparse}.
Finally, we briefly show the data collection criterion in Sec.~\ref{sec:collection}.

\subsection{Preliminary}
\label{sec:Preliminary}
Diffusion models~\cite{sohl2015deep,ho2020denoising,song2019generative,song2020score} were introduced as latent variable generative models with forward and reverse Markov chain, which gradually perturb the data with noise until tractable distribution and reverse the process with score matching or noise prediction for sampling.
Combined with prompt conditions, diffusion models are capable of generating images with aligned user aspiration.
In this work, we conduct experiments on the prevailing Stable Diffusion~\cite{rombach2022high}, which comprises an encoder $\mathcal{E}: \mathcal{X} \to \mathcal{Z}$, and a decoder $\mathcal{D}: \mathcal{Z} \to \mathcal{X}$, where $\tilde{\boldsymbol{x}} = \mathcal{D}(\mathcal{E}(\boldsymbol{x}))$. 
The denoising network $\epsilon_\theta$ is operated in the latent space with attached conditional encoder $\tau_\theta$. 
The training objective for stable diffusion is to minimize the denoising objective by
\begin{equation}
    \mathcal{L} = \mathbb{E}_{z,c,\epsilon,t}[\|\epsilon-\epsilon_{\theta}(\boldsymbol{z}_t,t,\tau_\theta(\boldsymbol{c}))\|^2_2],
    \label{eq:objective}
\end{equation}
where $\boldsymbol{z}_t$ is the latent feature at timestep $t$, and $\boldsymbol{c}$ is the given prompt condition.

Precedent customization methods involving instance-level optimization on specific subject with prior preservation loss~\cite{ruiz2023dreambooth,liu2023cones} or concept embedding~\cite{gal2022image,gu2024mix}. 
Recent practices on zero-shot customization largely derive from the stable diffusion and substitute the conditional encoder $\tau_\theta$ with image modal for semantic injection.
Moreover, the low-level signatures are complemented with paralleled model configurations such as ControlNet or ReferenceNet, which formulate the denoising network as $\epsilon_{\theta}(\boldsymbol{z}_t,t,\tau_\theta(\boldsymbol{c}),\zeta_\theta(\mathcal{T}(\boldsymbol{c})))$, where $\zeta_\theta$ provides the hierarchical subject interactions, and $\mathcal{T}$ is the transformation for delivering diversified hint signals~\cite{chen2023anydoor} with potential information bottleneck.

\vspace{-1em}
\begin{figure*}[h]
\begin{center}
   \includegraphics[width=1\linewidth]{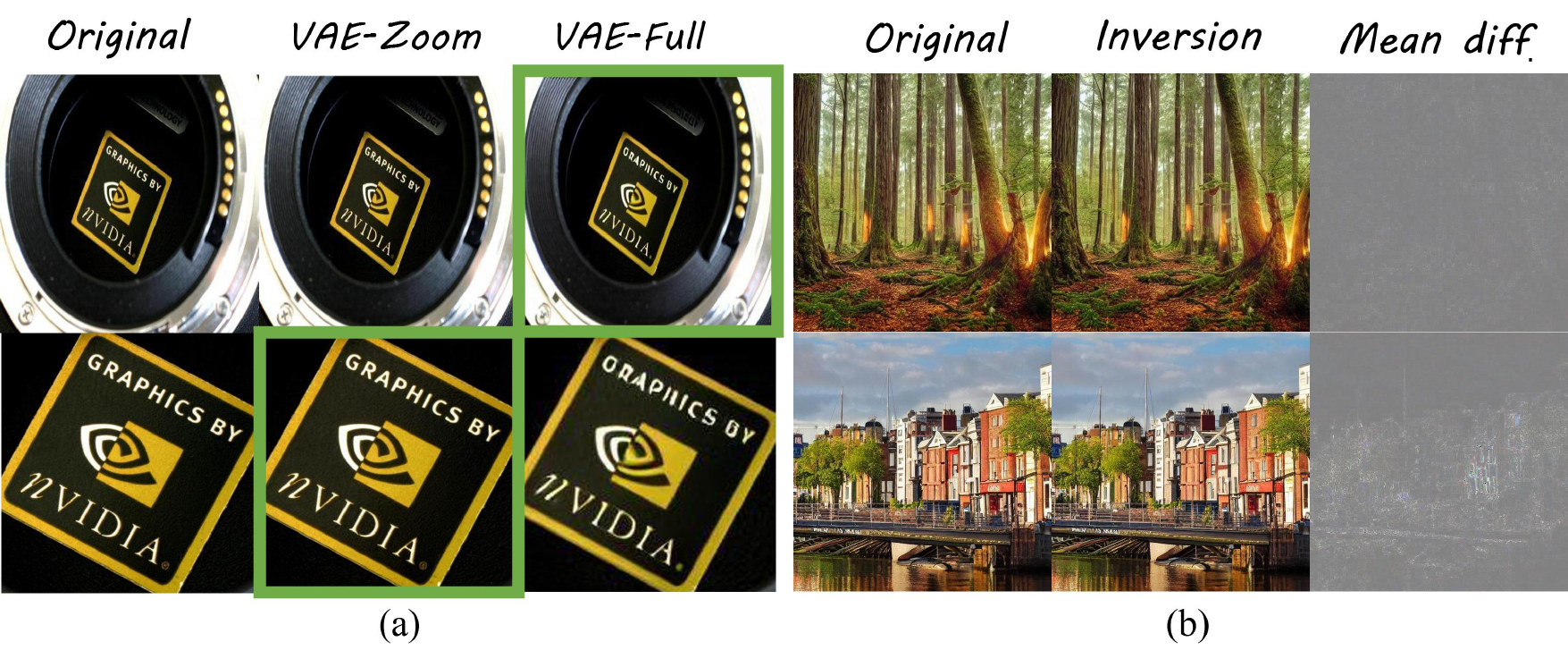}
\end{center}
\vspace{-1.6em}
   \caption{Preliminary of the informative latents in diffusion model. (a) Exclude the impact from the public VAE component, where the zoom-in reconstruction show almost spotless fidelity. (b) Without bells and whistles, the inversion technique reproduce the original image with single initial noise, the negligible deviation implying the great signature potential within the denoising model.}
\label{fig:preliminary}
\vspace{-1em}
\end{figure*}

\subsection{Symbiotic Diffusion System with Model Recycling}
\label{sec:recycling}
The intermediate latents in diffusion models contain rich semantic clues and copious fine-grained details~\cite{clark2024text,nguyen2024dataset,wang2023segrefiner}, while the derived downstream applications range from zero-shot classification, segmentation, to generative prompt-driven manipulation~\cite{tumanyan2023plug,liu2024towards}, and 3D rendering~\cite{karnewar2023holodiffusion,tang2023make}.
As shown in Fig.~\ref{fig:preliminary} (a), we first exclude the impact from the public VAE component, where the decoded vae latents with zoom-in transmission disclose the almost spotless fidelity, \ie, $\tilde{\boldsymbol{x}} \simeq \boldsymbol{x}$, implying the disengagement of the signature conservation.
Additionally, without bells and whistles, the glamorous images can be exactly generated through advanced diffusion inversion techniques~\cite{hong2023exact} with single initial noise, \ie, $\boldsymbol{x} \simeq \mathcal{D}(\Gamma(\boldsymbol{z}_t,t,\epsilon_\theta(\boldsymbol{z}_t,t,\tau_\theta(\boldsymbol{c}))))$, where $\Gamma$ is the noising schedule updated from $t: T\to0$, and $\boldsymbol{z}_T=\Gamma^{-1}(\boldsymbol{z}_t,t,\epsilon_\theta(\boldsymbol{z}_t,t,\tau_\theta(\boldsymbol{c})))$ obtained from $t: 0\to T$.
As shown in Fig.~\ref{fig:preliminary} (b), the appealing fidelity indicates the huge potential of the latent denoising model in signature collection, transmission, and representation along the denoising process.
Therefore, we suppose that \textit{the denoising network $\epsilon_\theta$ is sufficient for grasping the informative evidence we would desired about the signature}.
Disparate from prevailing approaches that dedicated to the external overconfigured assistant models $\zeta_\theta$, we made efforts toward the inside diffusion system for holistic efficiency.

\begin{figure*}[t]\small
\begin{center}
   \includegraphics[width=1\linewidth]{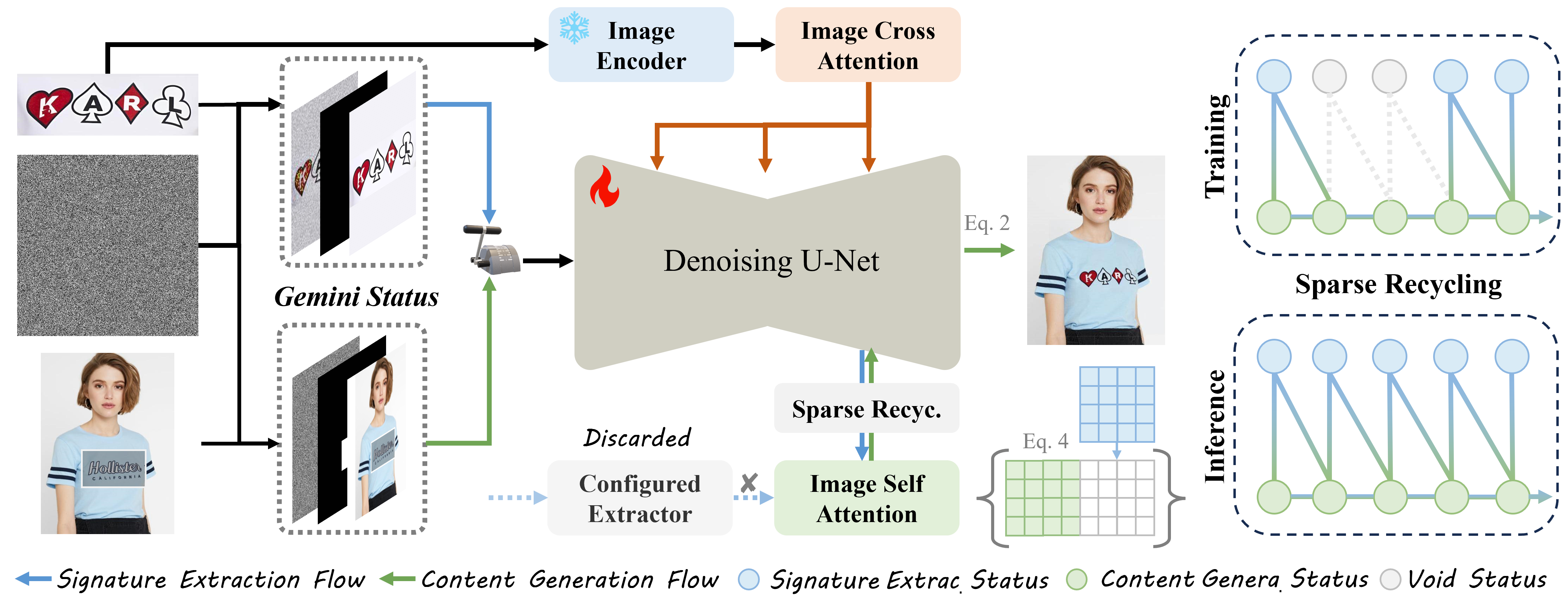}
\end{center}
\vspace{-1.6em}
   \caption{Overview of AnyLogo, which transports the customized textured subject to the candidate region in the scene image. The gemini status, \ie, signature extaction and content generation are performed alternatively in each denoising step. The overconfigured extractor is discarded with model recycling policy for signature delivery. The transmission quota is compressed during training, and released in inference for preventing the duplicate risk with steered diversified signature representation.}
\label{fig:method}
\end{figure*}

\textbf{Task Formulation.} We consider the subject-driven region customization, which ordinarily necessitate the scrupulous signature-level consistency, compared to the full-size image generation. 
The overall process involves the transportation of the arbitrary subject prompt $\boldsymbol{c}_{sub}$ to the candidate region specified by the binary mask $\mathcal{M}_{sce}$ of the scene image $\boldsymbol{x}_{sce}$ for seamless consolidation, formulated as
\begin{equation}
    \boldsymbol{x} = \Gamma(\hat{\mathcal{M}}_{sce} \odot \boldsymbol{z}_{sce}^{t} + (1-\hat{\mathcal{M}}_{sce}) \odot \boldsymbol{z}_{t},t,\epsilon_t)
    \label{eq:mask}
\end{equation}
where $\hat{\mathcal{M}}_{sce}$ is the interpolated mask formed to the latent size, $\boldsymbol{z}_{sce}^{t}$ is the encoded vae latent of $\boldsymbol{x}_{sce}$ with forward diffusion noise~\cite{ho2020denoising}, $\epsilon_t$ is the predicted $t$-step noise from the denoising network $\epsilon_\theta$.
The progressively infused scene latents provide the reliable background preservation.
In the following, we present the symbotic mechanism inside the $\epsilon_\theta$ for signature delivering.

\textbf{Recycling Policy.} The symbiotic diffusion system is built upon the model recycling policy with self-delivered signature payload. 
As shown in Fig.~\ref{fig:method}, the holistic subject-driven diffusion workflow is streamlined as vanilla image generation, where the accessorily configured consistency-relevant component $\zeta_\theta$ are discarded.
We discern that the rigorous signature collection and the creative content representation are promisingly compatible and can be systematically recycled within a single denoising network. 
Principally, the denoising objective of the symbiotic subject-driven diffusion system reinforces the vanilla image generation in Eq.~\ref{eq:objective}, and given by
\begin{equation}
    \mathcal{L} = \mathbb{E}_{z,c,\epsilon,t}[\|\epsilon-\epsilon_{\theta}(\hat{\boldsymbol{z}}_t,t,\tau_\theta(\boldsymbol{c}),\epsilon_\theta(\hat{\boldsymbol{z}}^{t}_{\mathcal{T}(\boldsymbol{c})}))\|^2_2],
    \label{eq:preobjective}
\end{equation}
where $\hat{\boldsymbol{z}}_t$ is the composition of the noisy latents $\boldsymbol{z}_t$, mask image latents $\hat{\mathcal{M}}_{sce} \odot \boldsymbol{z}_{sce}$, and binary mask $\hat{\mathcal{M}}_{sce}$, $\boldsymbol{z}^{t}_{\mathcal{T}(\boldsymbol{c})}$ is the encoded subject latent under the potential transformation $\mathcal{T}$ with $t$-step forward noise, and forming the  input space in the same way.
The signature extraction status is abbreviated from $\epsilon_\theta(\boldsymbol{z}^{t}_{\mathcal{T}(\boldsymbol{c})},t,\tau_\theta(\boldsymbol{c}))$ for simplicity, which shares the same workflow as content generation, except for the intermediate delivered signatures. 
We enforce the denoising objective solely in the content generation status, and released from the signature extraction, and the gemini status are alternate with concurrent timestep. 
Consistent with ~\cite{yang2023paint,tuo2023anytext,chen2023anydoor}, we replace the conditional encoder $\tau_\theta$ with image modal for semantic injection,
while the gemini status provide the hierarchical interactions inside the denoising network $\epsilon_\theta$ between the transmitted subjects $\boldsymbol{z}^{t}_{\mathcal{T}(\boldsymbol{c})}$ and generated contents $\boldsymbol{z}_t$. 
Explicitly, we cache the subject signatures at each self-attention procedure within the decoder part of the denoising network, and operate the inclusive mutual spatial attention in content generation flow. 
The delivered hierarchical signatures formulate the intersection of the gemini status, given by
\begin{align}
    \mathbf{C}^{t,i}_z = \text{Softmax} (\frac{Q^{t,i}_z\cdot \hat{K}^{t,i}}{\sqrt{d_{i}}})\cdot \hat{V}^{t,i},
\label{eq:softmax}
\end{align}
where $Q^{t,i}_z$ is the query and derived from the generated content latents $\boldsymbol{z}_t^{i}$ at $i$-th layer index, $\hat{K}^{t,i}$ and $\hat{V}^{t,i}$ are key and value and derived from the gemini status $[\boldsymbol{z}_t^{i},\boldsymbol{z}^{t,i}_{\mathcal{T}(\boldsymbol{c})}]$,
$d_i$ is the feature dimension, and $\mathbf{C}^{t,i}_z$ is the output of the current attention procedure.

\begin{figure*}[t]\small
\begin{center}
   \includegraphics[width=1\linewidth]{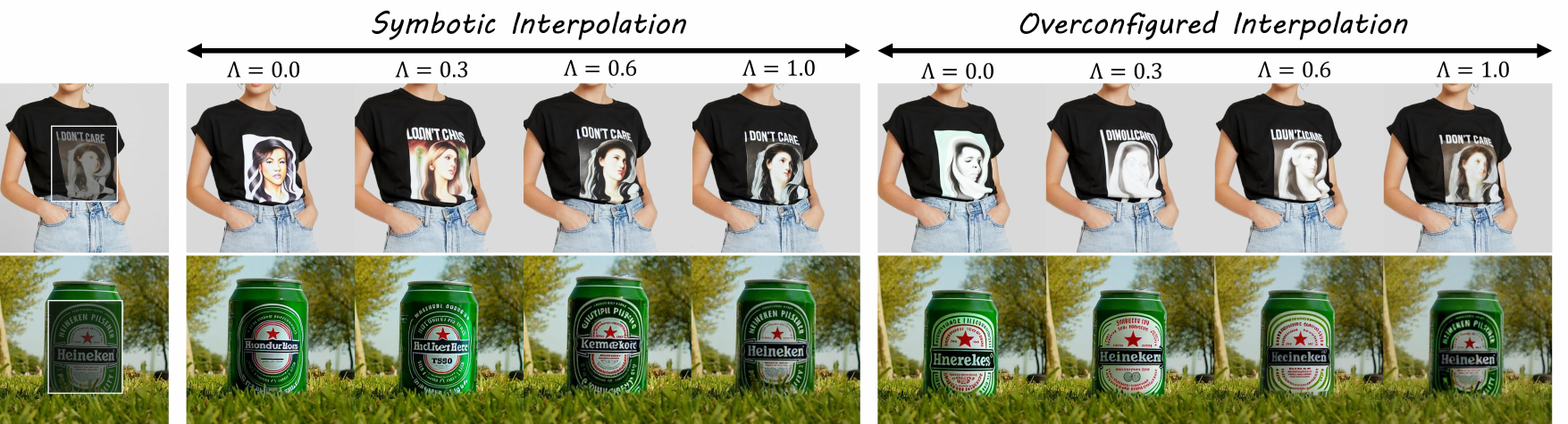}
\end{center}
\vspace{-1.6em}
   \caption{The comparison of the signature interpolation between the symbotic system and overconfigured system, where the signature flows are delivered progressively with increasing threshold, ranging from the blocked states to the fully released states during inference. The symbotic system manifests the consistent semantic content and comforting quality with flexible signature decoration.}
\label{fig:interpo}
\end{figure*}

\textbf{Symbiotic Temperament.} Peculiarly, the low-level signatures derived from the symbiotic diffusion system are highly semantic-independent, and the creative content generation capability is widely preserved even with the blocked signature flow $\epsilon_{\theta}(\boldsymbol{z}_t,t,\tau_\theta(\boldsymbol{c}),\varnothing)$, which is significantly different from the collapsed  quality and diversity in overconfigured systems, as shown in Fig.~\ref{fig:interpo}.
We suppose the auxiliary $\zeta_\theta$ induces the leakage of the generative expertise from the eantangled forward interdependence with denoising network $\epsilon_\theta$.
Moreover, we show that the symbiotic system enjoys the continuous transmission space for progressive signature decoration with controllable signature flow.

\subsection{Sparse Recycling}
\label{sec:sparse}
Albeit the scrupulous signature concentration, the overloaded transmission could cause the potential duplicate risk with ultimate subject fidelity and inferior scene conformity.
Prior investigations settle the information bottleneck with suppressive transformation $\mathcal{T}$ for signature delivering, including landmark representation and high-frequency map. 
However, the symbiotic diffusion system predefines the congruous status behavior, and requests the delivered subject within the same signal space as the denoising latents, \ie, the identity $\mathcal{T}$, resulting in complete signature disclosure.
In light of this, the sparse recycling paradigm is adopted for the symbiotic system with compressed transmission quota.
Sidestep the transformation of the input signal, we randomly discard the self-delivered signatures within the denoising network at cached attention procedures in the content generation flow.

\begin{wraptable}{t}{0.52\textwidth}
\vspace{-1.5em}
\Large
\caption{The transmission gain of the sparse recycling paradigm in delivering the faithful subject at both overconfigured system and symbotic system.}
\centering
\resizebox{0.52\textwidth}{!}{
    \begin{tabular}{lcccc}
    \toprule
    {\textbf{Configured}} & {CLIP-S $\uparrow$} & {LPIPS $\downarrow$}&{MUSIQ $\uparrow$}  \\
    \midrule
    ControlNet&87.9 (\textbf{+} 0.2)&0.141 (\textbf{+} 0.004)&68.6 (\textbf{+} 0.5)\\ 
    ReferenceNet&90.2 (\textbf{+} 0.4)&0.134 (\textbf{+} 0.011)&68.2 (\textbf{+} 0.6)\\
    \midrule
    Recycling&91.3 (\textbf{+} 0.3)& 0.127 (\textbf{+} 0.009)&68.9 (\textbf{+} 0.8)\\
    \bottomrule
    \end{tabular}}
\vspace{-1.5em}
\label{Tab:sparse}
\end{wraptable}

For consistency, we remain utilize $\mathcal{T}$ to denote the suppressive transformation, and given by
\begin{equation}
\mathcal{T}_{\epsilon_i,\Lambda}(\boldsymbol{c}) = 
\begin{cases}
\boldsymbol{z}^{i}_{\boldsymbol{c}}, & \text{if } k \leq \Lambda \text{ and } k \sim \mathcal{U}(0,1) \\
\varnothing, & \text{otherwise}
\end{cases}
\end{equation}
where $k$ is sampled from the uniform distribution $\mathcal{U}(0,1)$ for comparison with the threshold $\Lambda$, $\boldsymbol{z}^{i}_{\boldsymbol{c}}$ is the cached subject signature at $i$-th attention procedure, $\varnothing$ represents the null with discarding operation,
$\mathcal{T}_{\epsilon_i,\Lambda}$ is the layer-wise transformation attached to the denoising network $\epsilon_\theta$ that compress the signature flow. 
In particular, besides the prevention of the duplicated risk, the compressed transmission quota implicitly steers the diversified hyper-representations of the signature, facilitating the symbiotic system with improved subject transmission quality for scene harmonization,
as shown in Tab.~\ref{Tab:sparse}, which is also competent for the overconfigured system.
Consequently, the denoising objective of the symbiotic diffusion system with sparse recycling is given by
\begin{equation}
    \mathcal{L} = \mathbb{E}_{z,c,\epsilon,t}[\|\epsilon-\epsilon_{\theta}(\boldsymbol{z}_t,t,\tau_\theta(\boldsymbol{c}),\mathcal{T}_{\epsilon_\theta,\Lambda}(\boldsymbol{c}))\|^2_2],
    \label{eq:sparse_objective}
\end{equation}
where $\mathcal{T}_{\epsilon_\theta,\Lambda}$ is the collection of $\mathcal{T}_{\epsilon_i,\Lambda}$ along the denoising network. 
In inference, the sparse recycling is disabled in default with complete signature transmission, and enabled with continuous signature decoration.
Additionally, the classifier-free guidance~\cite{ho2021classifier} is incorporated to provide the conditional direction, formulated as 
\begin{equation}
\label{eq:cfg}
    \epsilon_{\theta,c}(\boldsymbol{z}_t,t,w,\boldsymbol{c}) =  
    \epsilon_\theta(\boldsymbol{z}_t,t,\varnothing,\varnothing) +  w(\epsilon_\theta(\boldsymbol{z}_t,t,\tau_\theta(\boldsymbol{c}),\mathcal{T}_{\epsilon_\theta,\Lambda}(\boldsymbol{c})) - \epsilon_\theta(\boldsymbol{z}_t,t,\varnothing,\varnothing)),
\end{equation}
where $w$ is the guidance scale parameter.

\subsection{Dataset Collection}
\label{sec:collection}
As there is a lack of publicly available dataset that tailored for the logo-level customization with rich graphic patterns, 
we present the BrushLogo-70k, collecting from the open data source comprises wild logo detection dataset such as OpenLogo~\cite{su2018open}, OSLD~\cite{bastan2019large}, virtual tryon dataset with brand region such as Dresscode~\cite{morelli2022dresscode}, VITON-HD~\cite{choi2021viton}, and text glyph dataset from AnyWord-3M~\cite{tuo2023anytext}.
The regions of interest are acquired with either ancillary provided or internal annotated, and the data are strictly excluded with irregular region size, aspect ratio, occlusion, and distortion quality. 
The detailed data composition is provided in the Appendix~\ref{sec: data-compos}.
The evaluation set is constructed with 1k high-quality entities extracted from the collection with upraised criterions, referred as AnyLogo-Benchmark.

\begin{figure*}[t]\small
\begin{center}
   \includegraphics[width=1\linewidth]{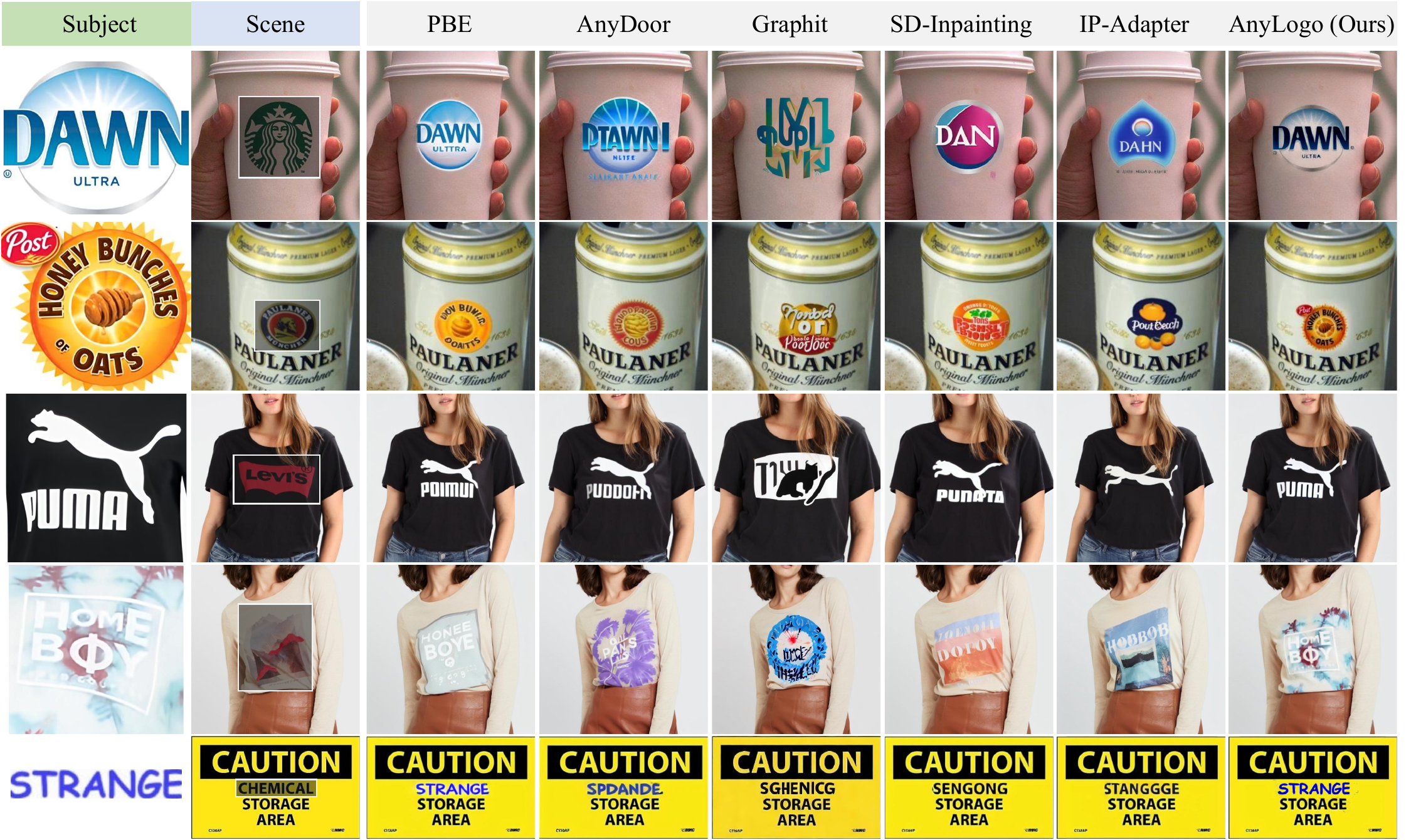}
\end{center}
\vspace{-1.6em}
   \caption{Qualitative comparison of AnyLogo and other competing methods on logo-level benchmark.} 
\label{fig:visual}
\end{figure*}

\begin{table}[t]
\renewcommand\arraystretch {1.3}
\caption{
Quantitative comparison with competing methods on logo-level benchmark. The subject fidelity are evaluated both from the semantic-level (CLIP-Score and DINO-Score) and signature-level (LPIPS), and the generated image quality is evaluated with MUSIQ.}
\Huge
\centering
\resizebox{1.0\textwidth}{!}{
    \begin{tabular}{lcccccccccccc}
    \toprule
    {} & \multicolumn{4}{c}{\textbf{Logo}} & \multicolumn{4}{c}{\textbf{Virtual Tryon}} &
    \multicolumn{4}{c}{\textbf{Text Glyph}}\\
    \cmidrule(lr){2-5}
    \cmidrule(lr){6-9}
    \cmidrule(lr){10-13}
    {\textbf{Method}} & {CLIP-S $\uparrow$} & {DINO-S $\uparrow$} & {LPIPS $\downarrow$} & {MUSIQ $\uparrow$} & {CLIP-S $\uparrow$} & {DINO-S $\uparrow$} & {LPIPS $\downarrow$} & {MUSIQ $\uparrow$} & {CLIP-S $\uparrow$} & {DINO-S $\uparrow$} & {LPIPS $\downarrow$} & {MUSIQ $\uparrow$} \\
    \midrule
    Paint-by-Example~\cite{yang2023paint}
    & 87.5 & 91.9 & 0.136 & 68.3 
    & 84.7 & 88.1 & 0.108 & 74.3 
    & 92.2 & 93.7 & 0.063 & 58.8 \\
    AnyDoor~\cite{chen2023anydoor}
    & 79.4 & 90.8 & 0.152 & 68.4 
    & 81.1 & 88.7 & 0.122 & 74.0 
    & 83.0 & 92.5 & 0.079 & 58.7 \\
    Graphit~\cite{graphit}
    & 67.7 & 85.4 & 0.195 & 67.4 
    & 68.9 & 80.3 & 0.182 & 73.7 
    & 75.4 & 88.5 & 0.112 & 58.2 \\
    SD-Inpainting~\cite{rombach2022high} 
    & 77.5 & 88.3 & 0.174 & 67.3 
    & 78.9 & 83.6 & 0.141 & 74.4 
    & 78.8 & 91.5 & 0.095 & 58.3 \\
    IP-Adapter~\cite{ye2023ip}
    & 82.7 & 89.9 & 0.148 & 68.0 
    & 82.1 & 84.9 & 0.125 & 73.9 
    & 83.4 & 91.2 & 0.083 & 58.7 \\
    \cmidrule(l){1-13}
    \rowcolor[HTML]{EFEFEF} AnyLogo (Ours)
    & \textbf{91.3} & \textbf{92.5} & \textbf{0.127} & \textbf{68.9} 
    & \textbf{91.4} & \textbf{94.0} & \textbf{0.082} & \textbf{74.6} 
    & \textbf{95.0} & \textbf{96.5} & \textbf{0.049} & \textbf{58.8} \\
    \bottomrule
    \end{tabular}}
\label{Tab:results}
\vspace{-0.2em}
\end{table}

\section{Experiments}
\label{sec: Experiments}
\subsection{Implementation Details}
\label{sec:implem}
We implement AnyLogo based on the Stable Diffusion v1.5 for weights initialization, while the VAE module is replaced with the SDXL version for advanced regression quality.
We train our model on constructed BrushLogo-70k dataset with 4 A800 GPUs for 50 000 steps.
We preprocess the scene images with zoom-in strategy, and the subject images are augmented with horizontal flip, rotation, optical distortion, and super resolution.
Image sizes are set to be 512×512.
We choose AdamW optimizer with the fixed learning rate of $1 \times 10^{-5}$ and the batch size of 64.
The weighted sampling strategy is adopted for unbalanced data composition.
The threshold in sparse recycling is set to be 0.6, 
and the conditional encoder is employed as DINOv2~\cite{oquab2023dinov2} for visual semantic injection.
During inference, we adopt the DDIM sampler with 20 denoising steps. More details in Appendix~\ref{sec: model-details}.

\textbf{Evaluation Metrics.}
The evaluation involves two aspects, the customized scenes region are supposed to be consistent with the provided subject, and the overall generated image should be photorealistic.
To this end, we introduce the following four metrics, 
the CLIP-score and DINO-score are adopted for measuring the subject fidelity from the profile-level with the cosine similarity between the extracted embeddings of the customized region and the subject, and the LPIPS~\cite{zhang2018unreasonable} is adopted for measuring the signature-level consistency.
The quality assessment metric  MUSIQ~\cite{ke2021musiq} is engaged to evaluate the authenticity and harmony of the overall generated image.
Additionally, following~\cite{ruiz2023dreambooth}, the diversified generation capability of the subject-driven diffusion system with blocked signature flow is further quantified with averaged LPIPS similarity between the generated images under the same subject.

\textbf{Baselines.}
We perform the comparison with following zero-shot region customization methods, including Paint-by-Example~\cite{yang2023paint}, AnyDoor~\cite{chen2023anydoor}, Graphit~\cite{graphit}, SD-Inpainting~\cite{rombach2022high}, and IP-Adapter~\cite{ye2023ip}.
SD-Inpainting is the text-driven method and we boost it with replaced CLIP image embeddings for subject transmission.
IP-Adapter is implemented with inpainting version.
The overconfigured system is implemented in the same experimental settings as AnyLogo for comparison, except for the extra configured ControlNet and ReferenceNet for signature delivering. 

\begin{figure}[t]
\centering
\vspace{-1em}
\begin{minipage}{0.54\linewidth}
\centering
\huge
\captionof{table}{Comparison of the overconfigured system and the symbotic system on wild logo customization.}
\renewcommand\arraystretch{1.3} {
\resizebox{\linewidth}{!}{
\begin{tabular}{lccccc}
\toprule[1.5pt]
Methods&CLIP-S$\uparrow$& DINO-S$\uparrow$& LPIPS$\downarrow$& MUSIQ$\uparrow$& DIV$\uparrow$ \\
\midrule
Baseline & 86.8 & 90.4& 0.137 & 68.5& -    \\
+ ControNet& 87.7 & 91.0 & 0.132 & 68.1& 0.213 \\
+ ReferenceNet& 89.8 & 91.3 & 0.123 & 67.6& 0.182 \\
\rowcolor[HTML]{EFEFEF} + \textbf{Model Recycling}& \textbf{91.0} & \textbf{92.1} & \textbf{0.118} & \textbf{68.1}& \textbf{0.279} \\
\bottomrule[1.5pt]
\end{tabular}}}	
\label{Tab:OS}
\end{minipage}
\hfill
\begin{minipage}{0.445\linewidth}
\centering
\huge
\captionof{table}{Ablation experiments of the recycling position in the denoising model for self-delivered signature transmission.}
\renewcommand\arraystretch{1.25} {
\resizebox{\linewidth}{!}{
\begin{tabular}{lcccc}
\toprule[1.5pt]
Position&CLIP-S$\uparrow$& DINO-S$\uparrow$& LPIPS$\downarrow$&MUSIQ$\uparrow$ \\
\midrule
Encoder & 89.2 & 90.5& 0.132  &67.5    \\
\rowcolor[HTML]{EFEFEF} Decoder& \textbf{91.0} & \textbf{92.1} & \textbf{0.118} & \textbf{68.1}\\
Enc. + Dec.& 90.3&91.7&0.127 & 67.7 \\
\bottomrule[1.5pt]
\end{tabular}}}	
\label{Tab:Pos}
\end{minipage}
\vspace{-0.6em}
\end{figure}

\begin{figure*}[t]\small
\begin{center}
   \includegraphics[width=1\linewidth]{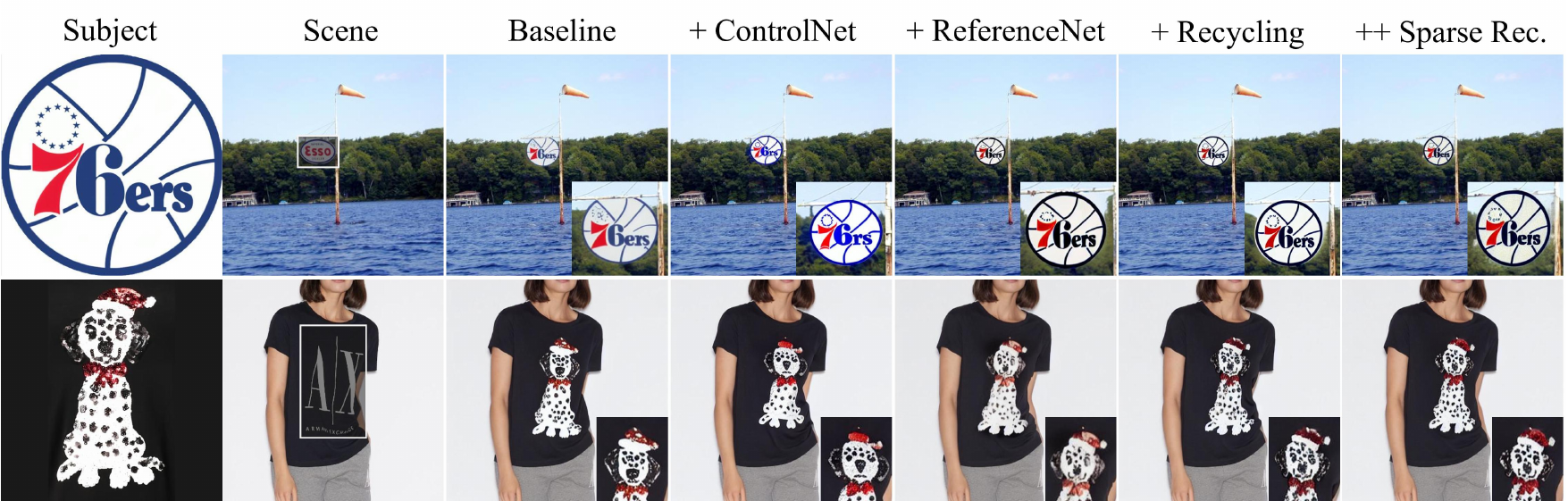}
\end{center}
\vspace{-1.2em}
   \caption{Visual comparisons of the overconfigured systems with configured ControlNet and ReferenceNet, and the symbotic system with model recycling policy and sparse transmission quota.} 
\label{fig:visual-abl}
\vspace{-1.2em}
\end{figure*}

\subsection{Comparison with Existing Alternatives}
We provide the quantitative comparison results in Tab.~\ref{Tab:results}, where the AnyLogo is superior in maintaining the signature consistency and semantic fidelity across diverse logo-level subjects, range from the wild brand logoes, Tryon patterns, and text glyphs. 
It can be observed that the solitary semantic injection is stumbling for signature-preserved customization.
Albeit the complementary hint signals strived by AnyDoor for signature transmission, the contour profile is informative insufficient with lossy compression, and more like spatial structure arrangement. 

The qualitative results are presented in Fig.~\ref{fig:visual}. Compared to the AnyLogo where the richly textured subjects are well transported to the candidate regions in the scene image with less distortion, other alternatives struggle in delivering the accurate low-level signatures with coarse semantic consistency, deviating from the color, pattern structure, and hallucination rendering.
It would be conscious that the signature-level consistency is dramatically differ from the object-level concentration, where the dispersed and disconnected subjects are toilsome to be grasped against the strongly semantic compact entity, and the interleaved text and graphic layout form the raised challenge. 

\subsection{Ablation Study}
\textbf{Overconfigured System.} 
We provide the comparison against the overconfigured system in Tab.~\ref{Tab:OS} with extra equipped ControlNet and ReferenceNet.
The baseline denotes the blocked signature flow with purely semantic injection.
It can be observed that albeit the improved faithfulness with extra configured signature extractor, the model recycling policy with self-delivered signatures achieves the excelling fidelity.
Note that the sparse recycling is excluded for validation.
The visual comparisons are provided in Fig.~\ref{fig:visual-abl}, where the transmission distortion of the overconfigured system are manifested as distorted color and structure compared to the symbotic system. 
We further evaluate the diversity of the two systems with disabled signature flow, where the symbotic system exhibit the higher diversity as quantified in Tab.~\ref{Tab:OS}, owing to the holistic system construction that preclude the entangled generative expertise leakage, and the visual comparisons are shown in Fig.~\ref{fig:interpo}.

\begin{wrapfigure}{r}{0.3\textwidth}
    \vspace{-.26in}
    \begin{center}
\includegraphics[width=0.3\textwidth]{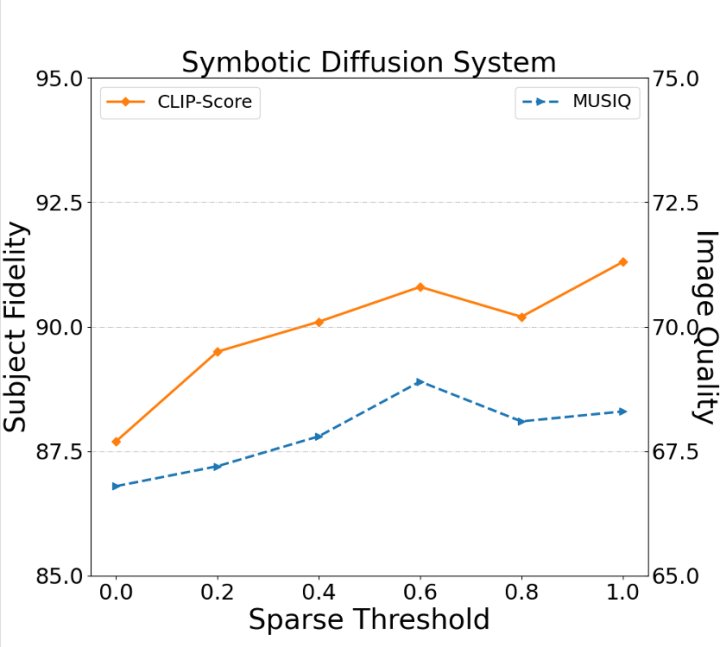}
    \end{center}
    \vspace{-.15in}
    \caption{Tendency of sparse recycling with progressive signature delivering thresholds.}
    \label{fig:sparse}
    \vspace{-.3in}
\end{wrapfigure}
\textbf{Position of the transmission.} In Tab.~\ref{Tab:Pos}, we provide the ablation experiments about the transmission position of the signature flow in model recycling policy.  
It can be observed that the transmission is efficient in the decoder part and encounters the obstacle in the encoder, which imply that the shallow layers are not well prepared with the steady semantic layout for signature delivering, and the overloaded signatures are undesirable during the content infancy. 

\textbf{Sparse Threshold $\Lambda$.} The sparse recycling paradigm is evaluated in Fig.~\ref{fig:sparse} with various transmission thresholds in optimization and fully released in inference.
We present the comparison curves both from the subject fidelity and image quality.
The uncompressed signature transmission induces the harmed image quality with disharmonious duplicated risk, also shown in Fig.~\ref{fig:visual-abl}.
The blocked signature flow manifests the restricted subject consistency with purely semantic injection.
Combined with sparse recycling, the symbotic system is much more efficient on semantic-relevant signature concentration with excluded signature noise.

\begin{figure*}[t]\small
\begin{center}
   \includegraphics[width=1\linewidth]{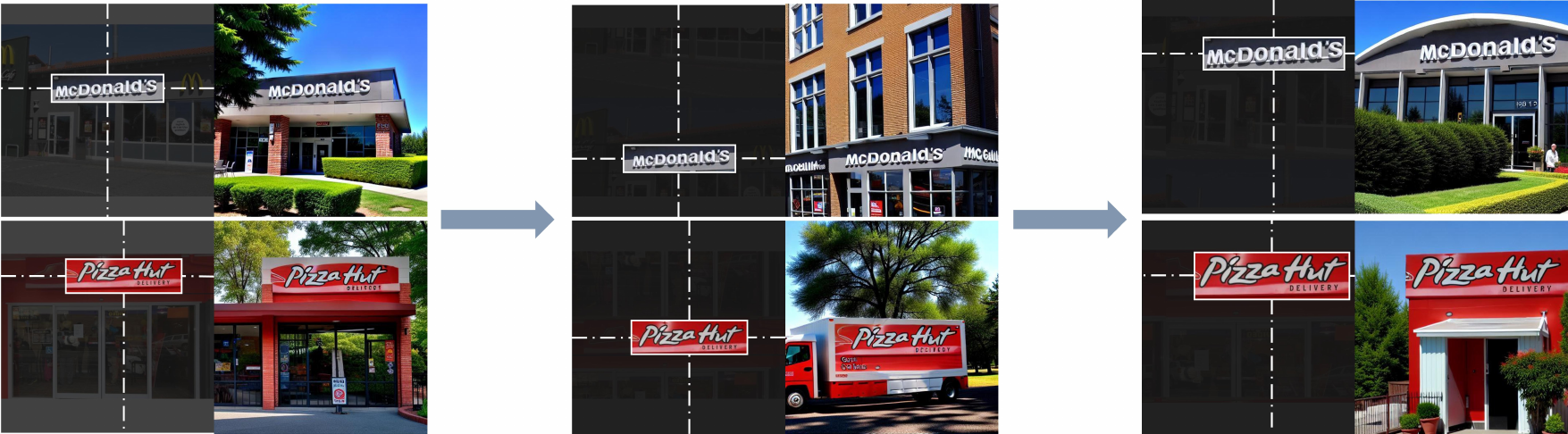}
\end{center}
\vspace{-1.3em}
   \caption{Outpainting results of AnyLogo, where the subject regions are preserved with absolute accuracy, and the scene background are regenerated with diversified presentation. The users are free to shift and scale the subject area for arbitrary highlighting display.}
\label{fig:outpaint}
\vspace{-1.2em}
\end{figure*}

\subsection{Discussion about the Fidelity and Future Works}
We show that AnyLogo is not only proficient in region customization with arbitrary user provided subjects, but also favour the diversified outpainting with faithful subject highlight, as shown in Fig.~\ref{fig:outpaint}, where the subject regions are preserved with absolute accuracy, and the scene background are regenerated with diversified presentation.
We are delighting to point that these are two ways to maintain the signature-level consistency.
In case of the lower fidelity requests for the scene region, 
it is of great potential to preserve the subject area for definitive fidelity with scale and shift manipulation, and the desired background could be complemented with semantic-faithful practices.
We note that the outpainting version of AnyLogo is effortless to be implemented with simply reversing the binary mask of the scene image both in input space and latents complementary.

\section{Conclusion}
In this work, we proposed AnyLogo, a symbiotic subject-driven diffusion system with remarkable low-level signature consistency.
Streamlined as vanilla image generation, we discern that the imperative customized gemini status, \ie, the rigorous signature extraction and creative content generation can be systematically recycled within a single denoising model and are promisingly compatible. 
The model recycling policy promotes the reinforced subject transmission efficiency with alleviated systematic coherence, and the disentangled semantic-signature space with continuous signature decoration.
Besides, the sparse recycling paradigm is adopted to prevent the potential duplicated risk with compressed transmission quota for diversified signature stimulation.
Extensive experiments on constructed AnyLogo-Benchmark demonstrate the effectiveness and practicability of our method.

\bibliographystyle{unsrt}
\bibliography{neurips_2024}

\medskip

\newpage
\appendix
\section{More Model Details}
\label{sec: model-details}
We implement the conditional encoder $\tau_\theta$ with hierarchical DINOv2 features for semantic injection, which conducted at all cross attention layers of the denoising model.
Specifically, we extract four group of embeddings from the DINOv2 that corresponding to the four scales of the denoising model.
And each embedding incorporates the spatial patch tokens with size of $\mathbb{R}^{257\times1024}$ and broadly dsitributed from the shallow to deeper layers with step size of 6. 
The extracted four group of semantic embeddings are progressively injected to the denoising model with symmetrical variation between encoder and decoder.
To perceive the background of the scene image for harmony subject transmission, we incorporates the additional background embedding that concatenates to the each group of subject embedding, which extracted in the same hierarchical manner, and each background embedding is only represented by single global token in size of $\mathbb{R}^{1\times1024}$ to exclude the scene details.

\section{Data Compositions}
\label{sec: data-compos}
The detailed data composition of the BrushLogo-70k and AnyLogo-Benchmark is illustrated in Tab~\ref{Tab:data}, which are adopted as train set and test set, respectively.
As presented in sec~\ref{sec:collection}, the wild logo are collected from the OpenLogo~\cite{su2018open} and OSLD~\cite{bastan2019large} with 23,947 and 25,955 subject-scene pairs, the brands in virtual tryon are collectd from the Dresscode~\cite{morelli2022dresscode} and VITON-HD~\cite{choi2021viton} with 5,434 and 2,986 pairs, and the text glyph are collected from the AnyWord-3M-Laion~\cite{tuo2023anytext} with 13,778 pairs.
\begin{wraptable}{t}{0.51\textwidth}
\renewcommand\arraystretch {1.1}
\vspace{-1.6em}
\Large
\caption{The composition of the collected dataset.}
\centering
\resizebox{0.51\textwidth}{!}{
    \begin{tabular}{lcccc}
    \toprule
    {\textbf{Data Split}} & Wild Logo & Virtual Tryon &Text Glyph \\
    \midrule
    BrushLogo-70k& 49382& 8420 & 13492 \\ 
    AnyLogo-Benchmark& 520 & 223 & 286 \\
    \bottomrule
    \end{tabular}}
\vspace{-1em}
\label{Tab:data}
\end{wraptable}
The test set is constrcuted with 1k high-quality pairs extracted from the aforementioned collections with upraised criterions.
The data are filtered with CLIP-IQA~\cite{wang2023exploring} from three dimensions with prompts of quality, contrast, and sharpness.
Apart from that, the irregular region size and aspect ratio are excluded.
The regions of the logo pattern or text in the scene images are acquired with either ancillary provided (wild logo and text glyph) or internal annotated (virtual tryon).
The subjects are paired with high DINO-simlarity from other entities under the same brand for the wild logo, and provided with paired product in the virtual tryon. 
The subjects for text are extracted from its own scene images to prevent glyph distortion.

\section{Limitations}
\label{sec: limitations}
\begin{wrapfigure}{r}{0.6\textwidth}
    \vspace{-.25in}
    \begin{center}
\includegraphics[width=0.6\textwidth]{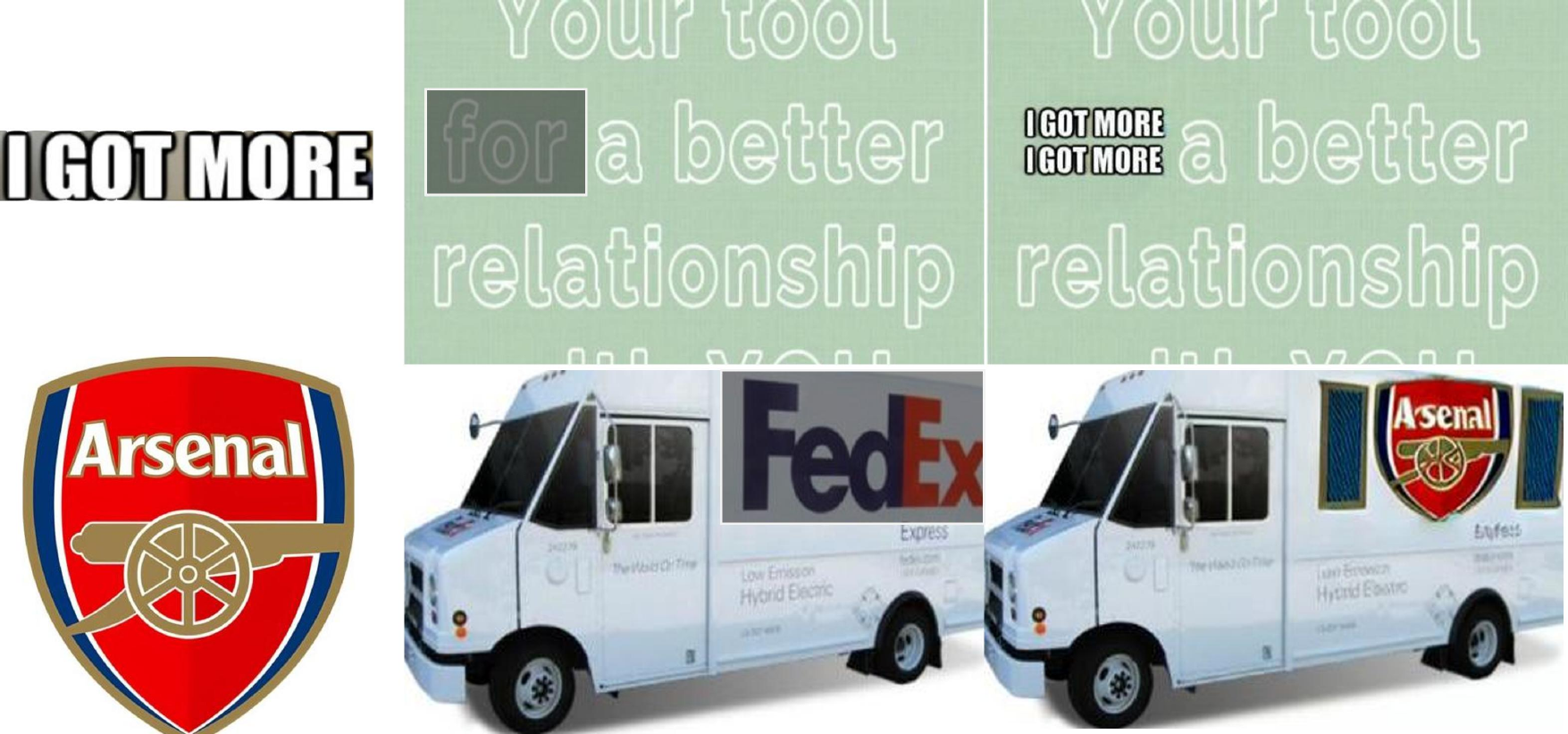}
    \end{center}
    \vspace{-.15in}
    \caption{The failure cases. The model tends to repeat the subject or hallucinate the new concept to fill the customized region when encounters the unmatched aspect ratio.}
    \label{fig:lim}
    \vspace{-.35in}
\end{wrapfigure}

The limitations are encountered with two main problems, as shown in Fig.~\ref{fig:lim}. The first is the unmatched aspect ratio between the subject and the customized region in scene image. In order to maintain the
subject fidelity with less distortion, we do not perform the resize operation in preprocessing, and the transmission are tend to repeat the subject with original ratio to fill the specified region.
The second is the hallucination that related to the scene image, which is also occurred with unmatched aspect ratio between the subject and the customized region. And the model tends to hallucinate the new concept that related to the background of the scene image, \eg,the  side windows for the car, to fill the candidate region.

\section{Boarder Impacts}
\label{sec: boarder}
The ability to manipulate logos could be benefit for product promotion, poster making, logo alteration in advertising position, \etc, and the outpainting version with diversified highlighting backgrounds could ease the cost of the venue rental and model hiring.
However, the misuse could incur the potential copyright problems with legal disputes.
And the generative logos may confuse the consumers to discern the authenticity and impact the reputation of the brands. 
Furthermore, the population of the generative models could impact the graphic designers and brand professionals, as the automated logo alteration with similar semantic layout might reduce the demand for manual design work.
We preclude the copyright infringement with infused watermarks~\cite{fernandez2023stable} to fingerprint the generated images.

\section{Comparison of the Overconfigured System and Symbotic System}
\label{sec: symbotic}
\begin{figure*}[t]\small
\begin{center}
   \includegraphics[width=1\linewidth]{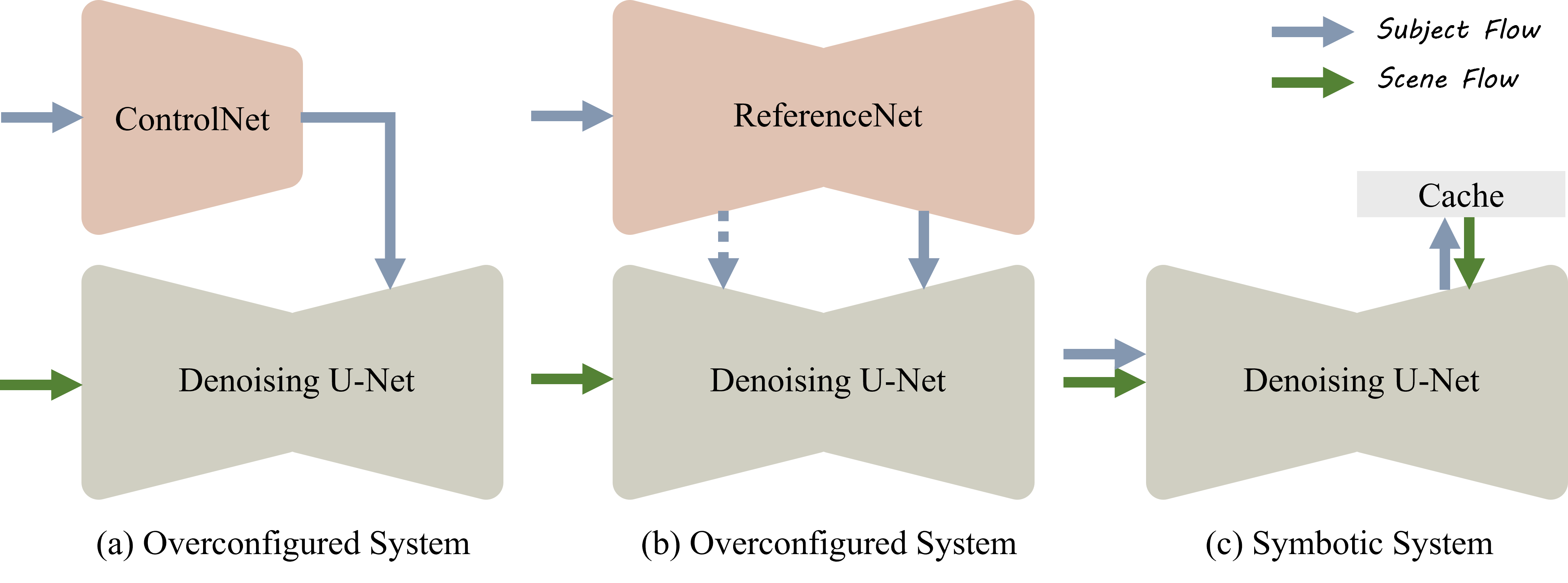}
\end{center}
\vspace{-1.6em}
   \caption{An illustration of the overconfigured diffusion system with extra configured ControlNet and ReferenceNet, and the symbotic diffusion system with model recycling, for subject customization.} 
\label{fig:symbotic}
\end{figure*}
We provide the detailed illustration of the overconfigured diffusion system and the symbotic diffusion system for subject customization in Fig.~\ref{fig:symbotic}.
As can bee seen that the overconfigured system employ the individual model to extract the signatures of the subject for reinforced detail enhancement, with progressive residual complement in ControlNet~\cite{tuo2023anytext,zhang2024brush,zang2024product,chen2023anydoor,seyfioglu2024diffuse}, and hierarchical spatial attention in ReferenceNet~\cite{hu2023animate,xu2023magicanimate,wang2024instantid,purushwalkam2024bootpig}. 
The symbotic diffusion system is built upon the model recycling policy with eliminated signature-relevant model configurations, and the signature extraction and content generation are systematically recycled within a single denoising model.

We provide the detailed calculation process of the Fig.~\ref{fig:stas}, where the transmitted statistic latent difference (SLD) is calculated with $\ell_1$ error of the average between the delivered subject latents and corresponding denoising latents in the delivered layer. 
We provide four layers comparison along the denoising model, including the middle-attention0, up1-attention2, up2-attention2, and up3-attention2, where the middle refers to the middle block of the denoising model, up refers to the decoder of the denoising model, and attention refers to the self-attention layer.
It can be observed that the overconfigured system exhibits the larger statistic discrepancy between the delivered subject latents and corresponding denoising latents, compared to the symbotic system, owing to the external engagement that breaks the statistic coherence.
Consequently, the accumulative subject attention (ASA) is significantly hampered, as shown in the shadow region of the Fig~\ref{fig:stas} (a), which calculated with proportion of the accumulated attention score of the subject against the overall attention map in the delivered self-attention layers

\section{More Visual Comparisons}
\label{sec: visual}
\begin{figure*}[h]\small
\begin{center}
   \includegraphics[width=1\linewidth]{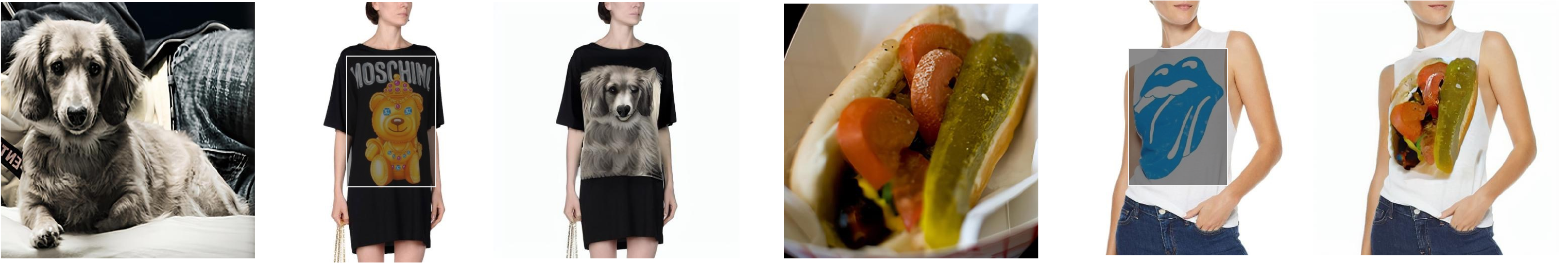}
\end{center}
   \caption{Visual results of AnyLogo on object customization.} 
\label{fig:object}
\end{figure*}

We provide the visual results of AnyLogo on object-level customization in Fig.~\ref{fig:object}, without tuning.
It can be observed that AnyLogo is proficient in rendering the portrait of the natural object for personali-

zation, which is applicable to customize the user assets with personalized preference, such as pet and favoured cuisine.

\begin{figure*}[t]\small
\begin{center}
   \includegraphics[width=1\linewidth]{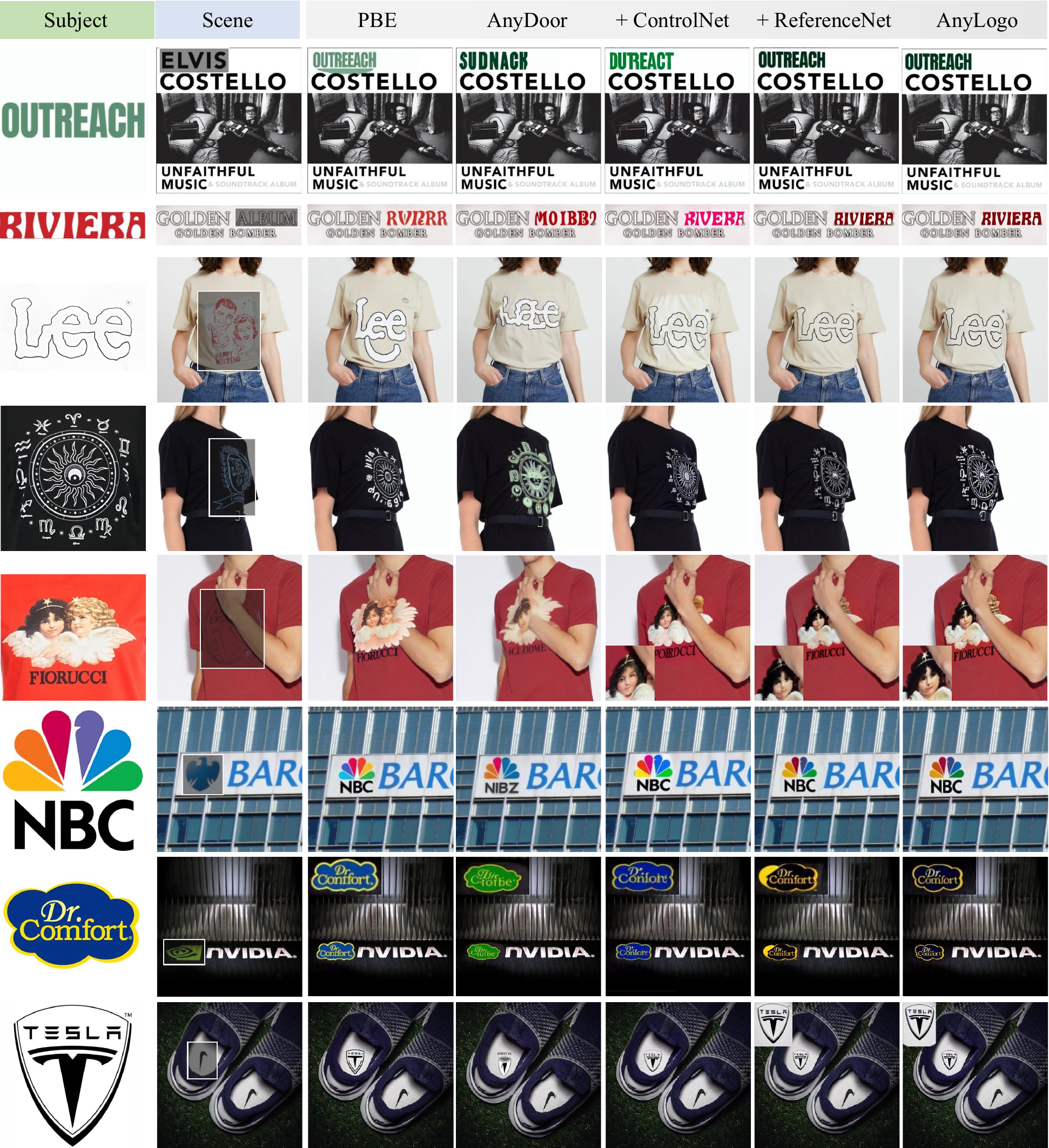}
\end{center}
   \caption{Visual comparison of AnyLogo and other competing methods on logo-level benchmark.} 
\label{fig:visual-app}
\end{figure*}
In Fig.~\ref{fig:visual-app}, we provide more visual comparison results of AnyLogo with other competing methods on logo-level customization, together with the overconfigured system equipped by ControlNet and ReferenceNet for visual ablation.
It can be observed that AnyLogo is proficient in maintaining the subject-specific signatures rather than hallucinating the general semantics, including curve contour, graphic layout, \etc, and dealing well with the scene background, such as occlusion.

\end{document}